\documentclass[10pt,twocolumn,letterpaper]{article}

\usepackage{cvpr}              
\usepackage[dvipsnames]{xcolor}

\definecolor{cvprblue}{rgb}{0.21,0.49,0.74}
\usepackage[pagebackref,breaklinks,colorlinks,citecolor=cvprblue]{hyperref}
\usepackage{multirow}
\def\paperID{3017} 
\def\confName{CVPR}
\def\confYear{2024}

\title{FreeDrag: Feature Dragging for Reliable Point-based Image Editing}

\author{Pengyang Ling$^{1*}$\quad
Lin Chen$^{1,2*}$\quad
Pan Zhang$^{2}$\quad
Huaian Chen$^{1\dag}$\quad 
Yi Jin$^{1\dag}$\quad 
Jinjin Zheng$^{1}$
\\
\textsuperscript{\rm 1}University of Science and Technology of China
\quad\quad
\textsuperscript{\rm 2}Shanghai AI Laboratory \\
{\tt\small \{lpyang27, chlin, anchen, jinyi08, jjzheng\}@mail.ustc.edu.cn}\quad
{\tt\small zhangpan@pjlab.org.cn} \quad
}

\begin{document}
\twocolumn[{%
\renewcommand\twocolumn[1][]{#1}%
\maketitle

\begin{center}
    \vspace{-8mm}
    \centering
    \captionsetup{type=figure}
    \includegraphics[width=0.99\textwidth]{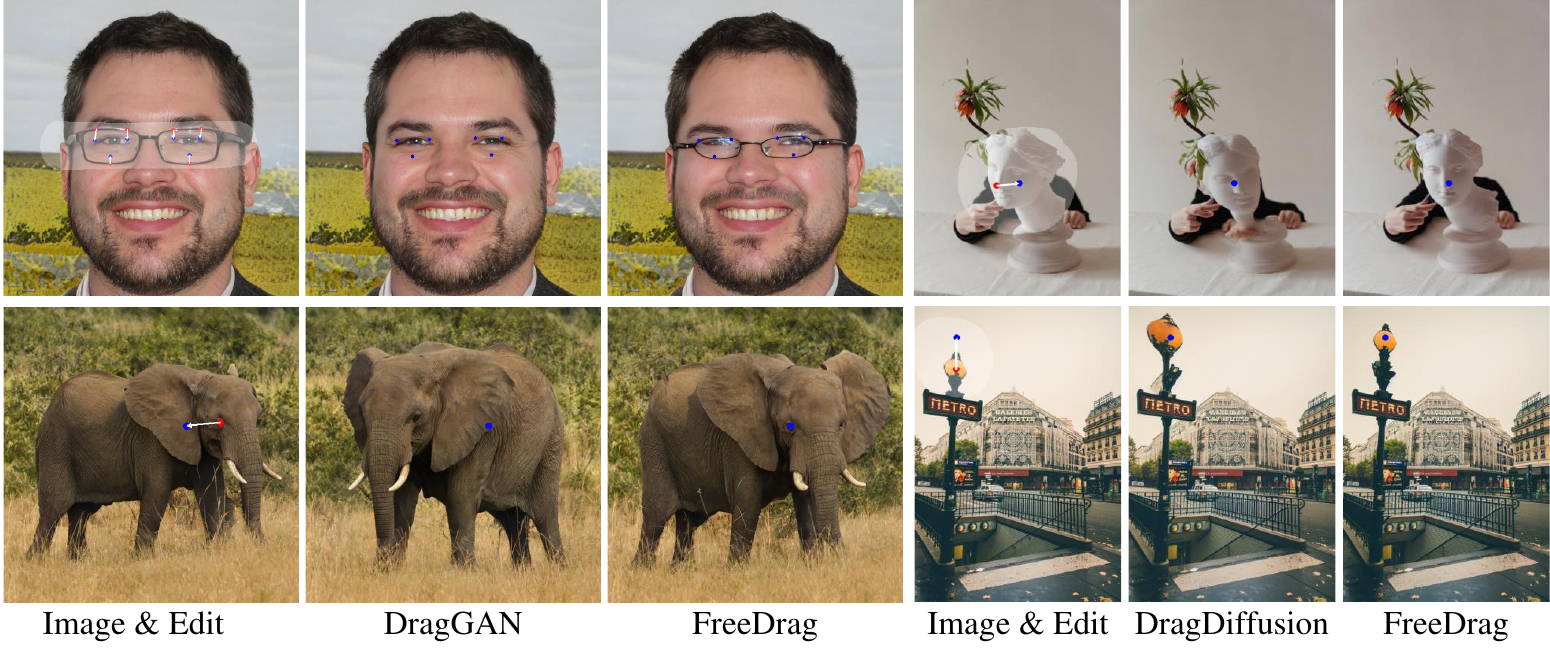}
    \vspace{-3mm}
    \captionof{figure}{ The comparison between the feature-centric FreeDrag and point-based DragGAN \cite{pan2023drag} and DragDiffusion\cite{dragdiffusion}.  Given an image input, users can assign handle points (\textcolor{red}{red} points) and target points (\textcolor{blue}{blue} points) to force the semantic positions of the handle points to reach corresponding target points, and optional mask can also be provided by users to assign editing region.  }
    \label{fig:figure-1}
    \vspace{-2mm}
\end{center}
}]
\renewcommand{\thefootnote}{\fnsymbol{footnote}}
\footnotetext[1]{~indicates equal contributions.}
\footnotetext[2]{~Corresponding authors.}

\begin{abstract}
To serve the intricate and varied demands of image editing, precise and flexible manipulation in image content is indispensable. 
Recently, Drag-based editing methods have gained impressive performance.
However, these methods predominantly center on point dragging, resulting in two noteworthy drawbacks, namely ``miss tracking", where difficulties arise in accurately tracking the predetermined handle points, and ``ambiguous tracking", where tracked points are potentially positioned in wrong regions that closely resemble the handle points. To address the above issues, we propose \textbf{FreeDrag}, a feature dragging methodology designed to free the burden on point tracking. The \textbf{FreeDrag} incorporates two key designs, i.e., template feature via adaptive updating and line search with backtracking, the former improves the stability against drastic content change by elaborately controlling the feature updating scale after each dragging, while the latter alleviates the misguidance from similar points by actively restricting the search area in a line. These two technologies together contribute to a more stable semantic dragging with higher efficiency. Comprehensive experimental results substantiate that our approach significantly outperforms pre-existing methodologies, offering reliable point-based editing even in various complex scenarios.
\end{abstract}    
\section{Introduction}
\label{sec:intro}

The domain of image editing utilizing generative models has gained substantial attention and witnessed remarkable advancements in recent years \cite{roich2022pivotal,parmar2023zero,mokady2023null,kawar2023imagic,hertz2022prompt,endo2022user}. In order to effectively address the intricate and diverse demands of image editing in real-world applications, it becomes imperative to achieve precise and flexible manipulation of image content. Consequently, researchers have proposed two primary categories of methodologies in this domain: (1) harnessing prior 3D models \cite{deng2020disentangled,ghosh2020gif,tewari2020stylerig} or manual annotations \cite{abdal2021styleflow,isola2017image,ling2021editgan,park2019semantic,shen2020interpreting} to enhance control over generative models, and (2) employing textual guidance for conditional generative models \cite{ramesh2022hierarchical,rombach2022high,saharia2022photorealistic}. Nevertheless, the former category of methodologies often encounters challenges in generalizing to novel assets, while the latter category exhibits limitations in terms of precision and flexibility when it comes to spatial attribute editing.

To tackle these aforementioned limitations, a recent pioneering study, known as DragGAN \cite{pan2023drag}, has emerged as a remarkable contribution in the realm of precise image editing. This work has garnered significant attention, primarily due to its interactive point-based editing capability, termed ``drag" editing, which enables users to exert precise control over the editing process by specifying pairs of handle and target points on the given image. The DragGAN framework introduces a two-step iterative process: (i) a motion supervision step, which directs the handle points to migrate towards their corresponding target positions, and (ii) a point tracking step, which consistently tracks the relocated handle points' positions. In each iteration, the points derived from the current iteration necessitate supervision from points of the last iteration and are subsequently tracked for the next iteration. We categorize this type of method, exemplified by DragGAN and its variant~\cite{dragdiffusion}, as point dragging solutions.

Notwithstanding the praiseworthy achievements exhibited by point dragging solution, there exist several issues. One issue is \textbf{miss tracking}, whereby point dragging encounters difficulty in effectively tracking the desired handle points. This issue arises particularly in highly curved regions with a large perceptual path length, as observed in latent space~\cite{stylegan-1}. In such cases, the optimized image undergoes drastic changes, leading to handle points in subsequent iterations being positioned outside the intended search region. 
Additionally, in certain scenarios, miss tracking leads to the disappearance of handle points, as shown in Figure \ref{fig:miss_tracking}. It is important to note that during miss tracking, the cumulative error in the motion supervision step increases progressively as iterations proceed, owing to the misalignment of tracked features. 
Another issue that arises is \textbf{ambiguous tracking}, where the tracked points are situated within other regions that bear resemblance to the handle points. This predicament emerges when there are areas in the image that possess similar features to the intended handle points, leading to ambiguity in the tracking process.
(see Figure \ref{fig:ambiguous_tracking}). This issue introduces a potential challenge as it can misguide the motion supervision process in subsequent iterations, leading to inaccurate or misleading directions.

To remedy the aforementioned issues, we propose \textbf{FreeDrag}, a feature dragging solution for interactive point-based image editing. To address the miss tracking issue, we introduce a template feature that is maintained for each handle point to supervise the movements during the iterative process. This template feature is implemented as an exponential moving average feature that dynamically adjusts its weights based on the errors encountered in each iteration. Even when miss tracking occurs in a specific iteration, the maintained template feature remains intact, preventing the optimized image from undergoing drastic changes. To handle the ambiguous tracking issue, we propose the line search with backtracking. Line search restricts the movements along a specific line connecting the original handle point and the corresponding target point. This constraint effectively reduces the presence of ambiguous points and minimizes the potential misguidance of the movement direction in subsequent iterations. Moreover, the backtracking mechanism enables prompt adjustment for motion plan by effectively discriminating abnormal motion, thereby enhancing the reliability of the total movement process. In light of the fact that the points in each iteration undergo supervision from template features and do not necessitate exacting tracking precision, we classify our approach as a feature dragging solution.
To summarize, our key contributions are as follows:

\begin{itemize}[leftmargin=10pt]
\itemsep0.4em
    
     
     \item We propose \textbf{FreeDrag}, a feature dragging solution for reliable point-based image editing that incorporates adaptive template features and line search with backtracking, marking a significant advancement in the field of flexible and precise image editing.
     
     \item We propose FreeDragBench, a new evaluation dataset with 2251 handmade dragging instructions that are tailored for GAN-based dragging editing, equipped with a new metric, which measures the editing accuracy of a pair of symmetrical dragging instructions.
     
\end{itemize}

\section{Related Work}

\subsection{Generative Adversarial Networks}
Generative adversarial networks (GANs)\cite{gan_goodfellow} have maintained the dominant position in image generation for an extended period. Classical unconditional GANs \cite{gan_review}, are devised to learn the mapping function from low-dimension random variables to realistic images that conform to the distribution of training datasets. Typically, the StyleGAN architecture \cite{stylegan-1,stylegan-2,stylegan_3,style_selfdistilled}, which employs a mapping network for low-dimension representation disentanglement and a synthesis network for photorealistic image generation, has made significant success in both generation quality and flexible style manipulation. Meanwhile, conditional GANs have been developed to enable versatile applications by infusing additional conditions, such as segmentation maps\cite{conditiongan_cut,conditiongan_seg_patchwise}, aerial photo\cite{conditiongan_aerial}, degraded images\cite{gan_enhancement,gan_derain,gan_dehaze}, and 3D variables \cite{conditiongan_3d_1,conditiongan_3d_2}.
\subsection{Diffusion Models}
The emerging diffusion models \cite{ddpm,ddim}, which conduct gradual denoising procedures from Gaussian noises to natural images, have recently sparked a strong wave of more potent image synthesis. Based on its promising generation capability, a series of versatile methods \cite{diffusionclip,diffusionblend,diffusioncontrol,diffusionholo3d,diffusionversatile} are developed to exceed the performance peaks of various generation tasks.
Typically, Rombach \etal propose the Latent Diffusion Model (LDM)\cite{stablediffusion}, which employs a pre-trained auto-encoder for perceptual compression and then performs high-quality sample in latent space, bringing a substantial advancement in high-resolution image synthesis. 
\subsection{Point-based Image Editing}
Given an image, interactive image editing aims to modify certain image content in response to specific user input, such as text instructions \cite{sdedit,deltaedit,instructpix2pix,sine}, region mask \cite{repaint}, and reference images \cite{paintbyexample, anydoor}.
The uniqueness of point-based image editing lies in that the user input is a set of point coordinates, and the generative models are expected to achieve precise image content manipulation to match the intent of users. For instance, Endo \cite{endo2022user} devises a latent transformer architecture to learn the mapping between two latent codes in StyleGAN. However, this framework necessitates the aid of a pre-trained optical flow network and demands a training procedure tailored for each model, which limits its practicability. Later, DragGAN \cite{pan2023drag} garners considerable attention with remarkable performance, which performs a cycle of point tracking and motion supervision in the feature map to persistently force the handle point to move to the target point. This simple framework achieves impressive performance and attracts subsequent works \cite{dragdiffusion, dragondiffusion} for better combination with the popular diffusion models. 

Generally, the GAN-based dragging approaches achieve superior dragging compared to diffusion-based approaches but exhibit inferior real image inversion. The GAN-based approaches benefit from the attribute disentanglement of StyleGAN, enhancing dragging capability.
However, its generative quality and real image inversion ability are comparatively limited.
In contrast, diffusion models achieve higher generative quality and superior real image inversion. Nevertheless, it encounters challenges in balancing point manipulation and appearance preservation due to the intertwined feature map, and demands more processing time. 

\section{Motivation}
\label{sec:motivation}

\begin{figure}[h]\centering
\vspace{-2mm}
\includegraphics[width=0.6\linewidth]{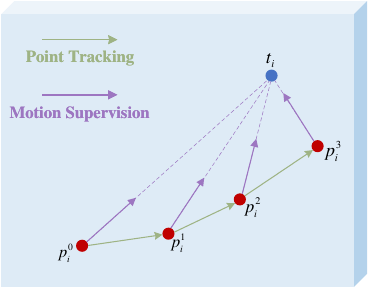}
\caption{Concept illustration of point dragging pipeline. $p_i^{k}$ denotes the tracked position of $i$-th handle point in $k$-th motion ($p_i^{0}=p_i$), and $t_i$ indicates the corresponding $i$-th target point.}
\label{fig:point_dragging}
\vspace{-6mm}
\end{figure}

Given a set of $n$ handle points $\left \{ p_1,p_2,p_3...,p_n \right \}$ and a corresponding set of $n$ target points $\left \{ t_1,t_2,t_3...,t_n \right \}$, the objective of point-based dragging is to displace $p_i$ to its respective $t_i$. Illustrated in Fig. \ref{fig:point_dragging}, the widely adopted DragGAN \cite{pan2023drag} accomplishes this objective through two sequential steps in each motion:
(i) Motion Supervision, wherein the current handle point is consistently directed towards its target point by leveraging the feature of itself.
(ii) Point Tracking, involving the search for the handle point in the proximity of the handle point from the last motion. Denoting the initial feature map as $F_0$, the tracked handle point $p_i^{k}$ for the $k$-th motion possesses the most similar feature to $F_0(p_i^{0})$ in the 2D tracking area centered at $p_i^{k-1}$.

\begin{figure}[h]\centering
\vspace{-2mm}
\includegraphics[width=0.95\linewidth]{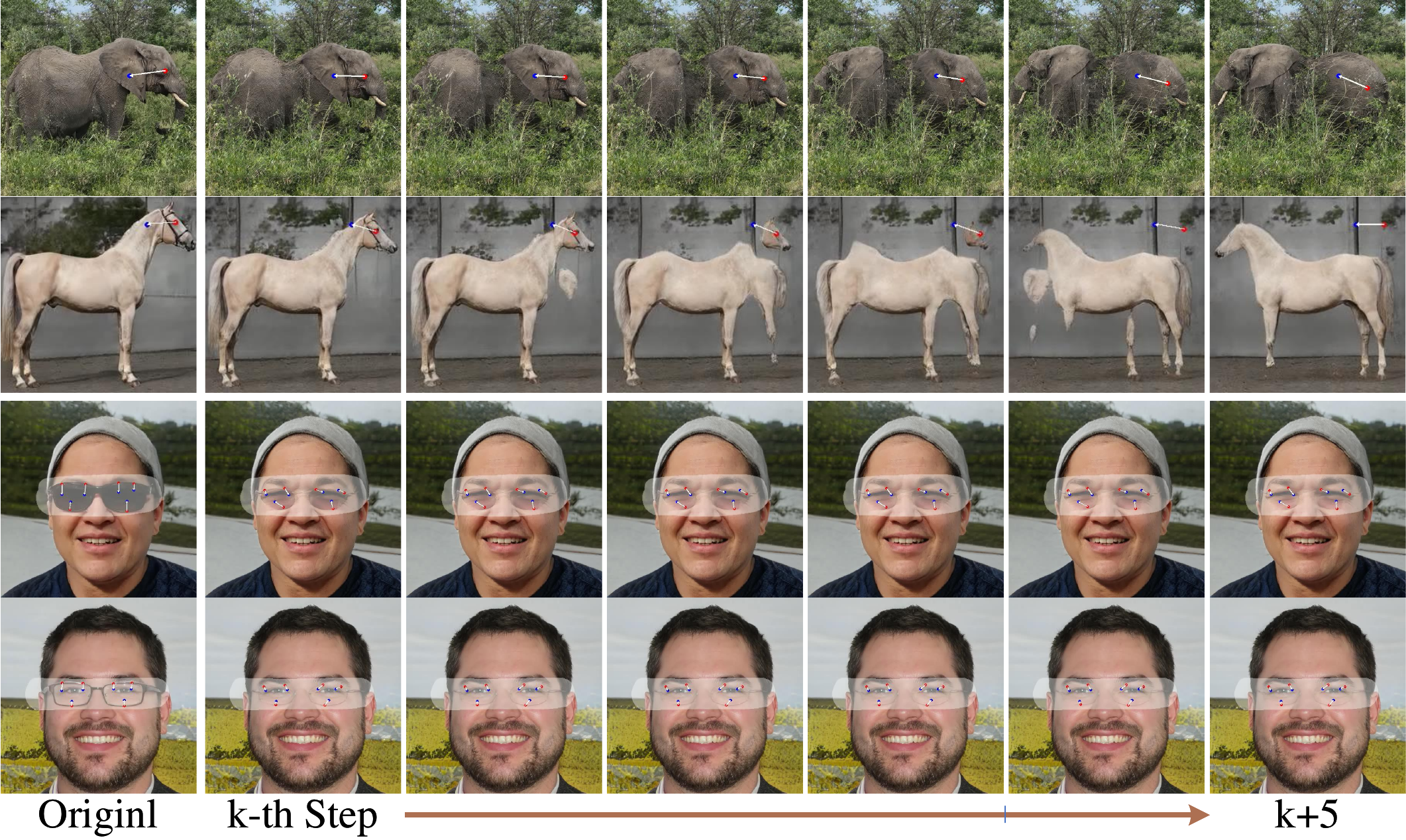}
\vspace{-2mm}
\caption{\textbf{Miss tracking} of DragGAN \cite{pan2023drag} due to the drastic change in layout (first and second rows) and the disappearance of handle points (third and last rows).}
\label{fig:miss_tracking}
\vspace{-2mm}
\end{figure}

\begin{figure}[h]\centering
\vspace{-2mm}
\includegraphics[width=0.85\linewidth]{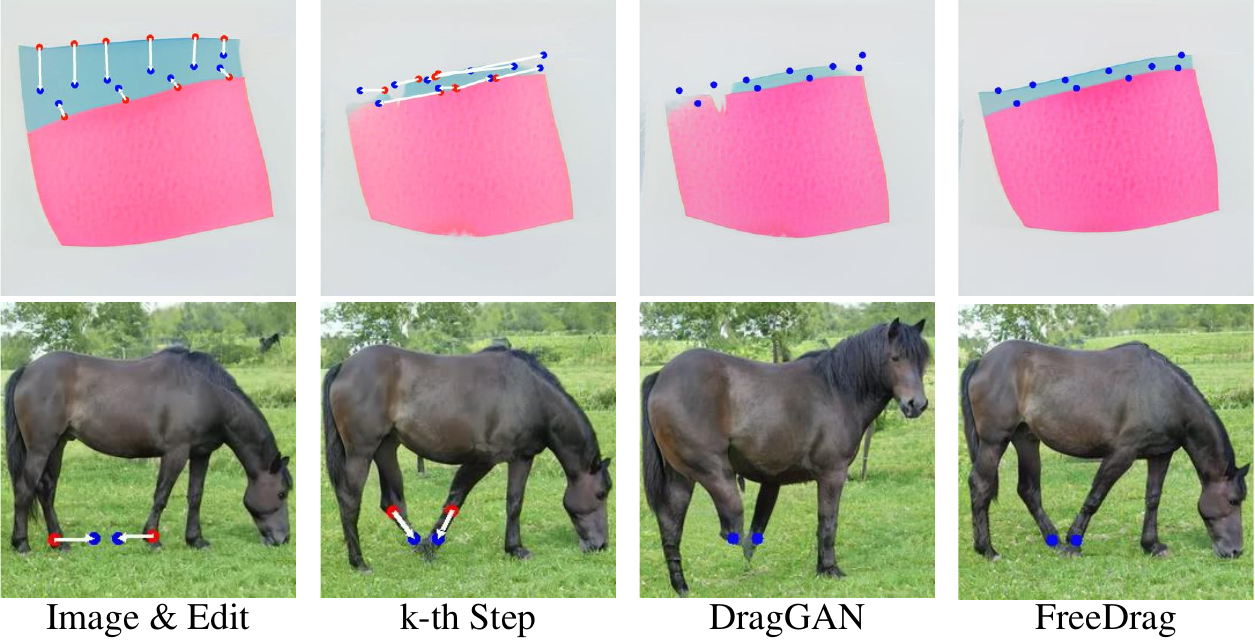}
\vspace{-2mm}
\caption{\textbf{Ambiguous tracking} in DragGAN \cite{pan2023drag} due to the existence of similar structures.}
\label{fig:ambiguous_tracking}
\vspace{-2mm}
\end{figure}

\begin{figure*}[h]\centering
\includegraphics[width=0.9\linewidth]{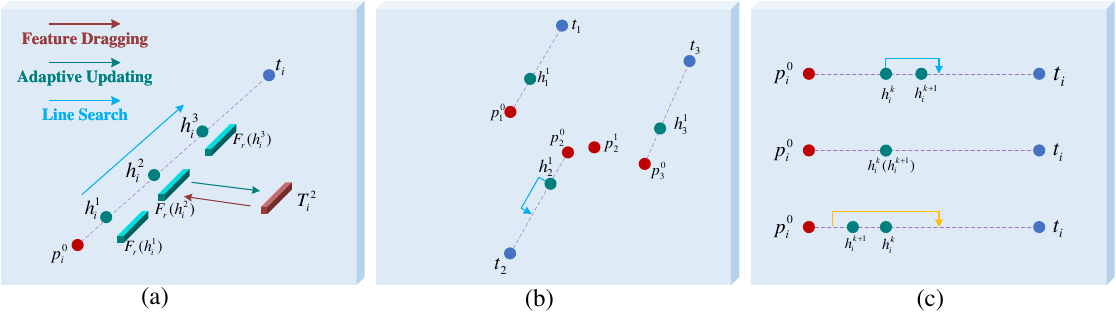}
\vspace{-3mm}
\caption{Illustration of proposed feature dragging pipeline. $h_i^{k}$ denotes the searched point in $k$-th drag, which lies in the line formed by $p_i^{0}$ and $t_i$, and $T_i^{k}$ denotes the corresponding template feature. (a) Concept of feature dragging. (b) The coupling movement under multiple points dragging. (c) The visualization of Eq. \ref{eq.whole_localization}. }
\vspace{-2mm}
\label{fig:feature_dragging}
\vspace{-3mm}
\end{figure*}

While the point dragging pipeline depicted in Fig. \ref{fig:point_dragging} presents a promising solution for point-based image editing, it is noted that it frequently encounters challenges, including handle point loss, imprecise editing, and distorted image generation in certain scenarios. These issues are attributed to the intrinsic instability of point dragging, encompassing miss tracking and ambiguous tracking. 
(i) Miss Tracking: This occurs in situations where point dragging encounters difficulty in effectively tracking the designated handle points. Given the presence of highly curved regions with substantial perceptual path lengths, as discerned in latent space~\cite{stylegan-1}, the optimized image undergoes significant alterations following motion supervision. Consequently, the handle point $p_i^{k+1}$ deviates outside the intended search region of $p_i^k$, as shown in Figure \ref{fig:miss_tracking}, leading to miss tracking in the point tracking step. In specific scenarios, $p_i^{k+1}$ may completely vanish from the entire feature map, exemplified by the disappeared glasses in Figure \ref{fig:miss_tracking}. It is imperative to underscore that during miss tracking, the cumulative error in the motion supervision step progressively amplifies with iterations due to the misalignment of tracked features.
(ii) Ambiguous Tracking: This occurs when the tracked points are positioned within other regions that bear resemblance to the handle points. This challenge arises when there are areas in the image exhibiting features similar to the intended handle points, such as the blue boundary lines and horse's hooves in Figure \ref{fig:ambiguous_tracking}, which may misdirect the motion supervision process in subsequent iterations, resulting in inaccurate or misleading directional adjustments.

\section{Methodology}
\label{sec:methodology}

In light of the instability associated with point dragging, which heavily depends on accurate point tracking in each step, we introduce a feature dragging approach termed \textbf{FreeDrag}, as illustrated in Fig. \ref{fig:feature_dragging}(a). Here, $h_{i}^{k}$ represents the target position in the $k$-th drag, and $F_{r}(h_i^{k})$ signifies the feature aggregate centered at $h_i^{k}$ with a radius $r$ in the feature map $F$, which can be expressed as:
\vspace{-3mm}
\begin{equation} \label{feature_aggregate} 
    F_r({h_i^{k}}) = \sum_{q_i\in \Omega (h_i^{k}, r)}F(q_i).
    \vspace{-3mm}
\end{equation}
Here, $\Omega (h_i^{k}, r)$ denotes the square patch centered at $h_i^{k}$ with a side length of $2r$. In the $k$-th drag, we promote $h_i^{k}$ to be the carrier of $T_i^{k}$ by compelling the feature aggregate $F_r(h_i^{k})$ to closely align with the template feature $T_i^{k}$ (as depicted by the \textbf{red line} in Fig. \ref{fig:feature_dragging}(a)), \textit{i.e.}, 
\vspace{-3mm}
\begin{equation}\label{eq.drag_supervision}
  {\mathcal{L}_{drag}} =  \sum_{i=1}^{n} {\left\| F_r(h_i^{k}) - T_i^{k}   \right\|_1}.
  \vspace{-2mm}
\end{equation} 

In order to facilitate high-quality feature dragging, multiple optimization steps are performed from the same position $h_i^k$, with consistent supervision as defined in Eq. \ref{eq.drag_supervision}.

The template feature undergoes adaptive updating according the quality of each dragging, as detailed in Section \ref{subsection:template_feature}. This updated template feature guides the feature of the handle point in the subsequent dragging. By gradually compelling $h_i^{k}$ to approach $t_i$, the template feature effectively transitions to the final $t_i$, indirectly encouraging the handle point to move towards the ultimate position. Additionally, we enforce constraints on $h_i^{k}$ and iterate to update the subsequent handle point $h_i^{k+1}$ along the line extending from $p_i^{0}$ to $t_i$ (as illustrated by the \textbf{blue line} in Fig. \ref{fig:feature_dragging}(a)). This approach not only provides a reliable movement direction but also significantly reduces the risk of misguidance arising from potential similar points.

\subsection{Template Features via Adaptive Updating}
\label{subsection:template_feature}

Concerning the template feature, it necessitates retaining the feature of the initial handle point on one hand, while on the other hand, it should undergo updates to accommodate reasonable geometric and appearance changes in each dragging. Accordingly, we introduce an adaptive updating strategy that permits a flexible updating scale, enabling the template feature to undergo few updates in chaotic situations and more updates in fine conditions. Specifically, the adaptive updating strategy for the template feature is formulated as follows:
\vspace{-2mm}
\begin{equation} \label{eq:adaptive_updating}
    T_{i}^{k+1} = \lambda_{i}^k   \cdot {F_r}(h_{i}^k) + (1 - \lambda_{i}^k  ) \cdot T_{i}^{k}.
\vspace{-2mm}
\end{equation}
Here, $\lambda_{i}^{k}$ represents the coefficient controlling the updating scale of the template feature in the $k$-th dragging. For consistency, we specifically define $\lambda_{i}^0 = 0$, $h_{i}^0 = p_{i}^0$, and $T_{i}^{0} = F_r(p_{i}^{0})$.    
Intuitively, for the $k$-th dragging, a smaller $\lambda_{i}^{k}$ is employed for low-quality feature dragging. This aids in maintaining $T_{i}^{k+1}$ relatively constant in chaotic situations. Conversely, a larger $\lambda_{i}^{k}$ is utilized for high-quality feature dragging, promoting sufficient updating of $T_{i}^{k+1}$ in fine conditions.

For simplicity, the feature discrepancy of between ${F_r}(h_{i}^k) $ and $T_{i}^{k}$ is denoted as $L_{(i,k)}$. Since Eq. \ref{eq.drag_supervision} is reused in multiple optimization steps for each feature dragging, we define $L_{(i,k)}$ at the initial/end optimization step in each dragging as $L_{(i,k)}^{in}$ and $L_{(i,k)}^{en}$, respectively. Accordingly. $L_{(i,k)}^{in}$ controls the difficulty of $k$-th feature dragging from $T_{i}^{k}$ to ${F_r}(h_{i}^k) $, and a larger $L_{(i,k)}^{in}$ indicates more arduous challenge for feature dragging.  While $L_{(i,k)}^{en}$ reflects the quality of each feature dragging, i,e, a smaller $L_{(i,k)}^{en}$ means fewer discrepancy between $T_{i}^{k}$ and ${F_r}(h_{i}^k) $ at the last optimization step, which implies higher quality feature dragging from $T_{i}^{k}$ to ${F_r}(h_{i}^k) $. Therefore, the adaptive coefficient $\lambda_i^{k} $ in Eq. \ref{eq:adaptive_updating} is devised as:
\vspace{-2mm}
\begin{equation}
    \lambda_i^{k}  = {(1 + exp(\alpha  \cdot (L_{(i,k)}^{en}- \beta )))^{ - 1}},
\vspace{-2mm}
\end{equation}
where $\alpha$ and $\beta$ denote two positive constants, and $exp(\cdot)$ represents the exponential function. Given a hyperparameter $l$, we determine $\alpha$ and $\beta$ by considering the following typical scenarios: (i) the well-optimized case, where we set $L_{(i,k)}^{en} = 0.2 \cdot l$ with $\lambda=0.5$; and (ii) the ill-optimized case, where we set $L_{(i,k)}^{en} = 0.8 \cdot l$ with $\lambda=0.1$, \textit{i.e.},
\vspace{-2mm}
\begin{equation}
0.5 = {(1 + exp(\alpha  \cdot(0.2 \cdot l - \beta )))^{ - 1}},
\vspace{-1mm}
\end{equation}
\begin{equation}
0.1 = {(1 +  exp( \alpha  \cdot(0.8 \cdot l - \beta )))^{ - 1}}.
\vspace{-2mm}
\end{equation}
Solving the equation yields $\alpha = \ln(9)/(0.6 \cdot l)$ and $\beta = 0.2 \cdot l$. It is noteworthy that we impose a constraint on the maximum value of $\lambda$ to mitigate the potential impact of incorrect updating.

\subsection{Line Search with Backtracking}
\label{subsection:line_search}
For the target position $h_i^{k}$ in the $k$-th dragging, we contemplate its localization from two perspectives: i) Reliable motion direction; ii) Appropriate feature discrepancy at the beginning of each drag, denoted as $L_{(i,k)}^{in}$. A too small value of $L_{(i,k)}^{in}$ fails to furnish adequate discrepancy in Eq. \ref{eq.drag_supervision} for gradient optimization, while an excessively large $L_{(i,k)}^{in}$ heightens the risk of unsuccessful feature dragging.

From the first goal, illustrated in Fig. \ref{fig:feature_dragging}(a), we constraint $h_i^{k}$ to the line extending from $p_i^{0}$ to $t_i$. This approach not only ensures a reliable movement direction but also markedly diminishes the risk of misguidance stemming from potential similar points. As for the second goal, point localization is conducted based on both feature discrepancy and motion distance, expressed as:
\begin{align}\label{eq.localization}
    h_i^{k+1} &= {S}(h_i^{k},t_i, T_i^{k+1},d,l) \\ 
      &= \mathop {\arg \min}\limits_{{q_i \in  \pi (h_i^k, t_i, d)}}\left\lVert \lVert F_r(q_i) - T_{i}^{k+1} \rVert_1 - l \right\rVert_1,
\end{align}
where $l$ and $d$ are two hyperparameters that control initial feature distance $L_{(i,k)}^{in}$ and maximum single movement distance, respectively, and $\pi (h_i^k, t_i, d)$ represents the point set, which includes $h_i^{k} + j\cdot \frac{{{t_i} - h_i^k}}{{{{\left| {{t_i} - h_i^k} \right|}_2}}}$ with $j= 0.1\cdot d, 0.2\cdot d,...,d$.

Additionally, as depicted in Fig. \ref{fig:feature_dragging}(b), during the joint optimization of multiple points dragging, the motion direction of a specific point may be influenced by the overall trend. This can result in the handle point deviating from the target point in certain steps. For instance, in comparison to $p_2^0$, the handle point $p_2^1$ is farther away from $h_2^1$. To address this issue, we integrate a backtracking mechanism to identify such abnormal movements, facilitating prompt adjustments for the subsequent dragging plan. Concretely, backtracking is implemented by introducing two additional options for the dragging plan: frozen and fallback the point, which can be expressed as:
\vspace{-2mm}
\begin{equation}\label{eq.whole_localization}
    h_i^{k + 1}\! =\!\! \left\{ \! \! \begin{array}{l}
{S}(h_i^{k},t_i, T_i^{k+1},d,l),
\quad if \ L_{(i,k)}^{en} \le 0.5\cdot{l}\\
\ \ \ \ \ \  h_i^k,\ \ \ \ \ \ \ \ \ \   \ \ \ \ \quad \ elif  \ \ \  L_{(i,k)}^{en} \le L_{(i,k)}^{in}\\
S(h_i^k \!-\! d \cdot \frac{{{t_i} - h_i^k}}{{{{\left\| {{t_i} - h_i^k} \right\|}_2}}},t_i,T_i^{k+1}, 2d,0), else
\end{array} \right.
\vspace{-2mm}
\end{equation}
 
For better comprehension, Eq. \ref{eq.whole_localization} has been visually represented in Fig. \ref{fig:feature_dragging}(c). To elaborate, the first scenario corresponds to a normal high-quality optimization, where $h_i^{k+1}$ closer to $t_i$ is assigned for further movement (depicted by the blue line in Fig. \ref{fig:feature_dragging}(c)). The second scenario corresponds to insufficient feature dragging, where $h_i^{k}$ is reused as $t_i^{k+1}$ for continued feature dragging towards the same point. In the exceptional case, \textit{i.e.}, $L_{(i,k)}^{en} > max\left \{ 0.5\cdot l,L_{(i,k)}^{in} \right \} $, we set $l=0$ and double the search range (illustrated by the yellow line in Fig. \ref{fig:feature_dragging}(c)) to immediately locate the point closest to the template feature $T_i^{k+1}$, promptly avoiding deterioration.

\subsection{Termination Signal}
For each feature dragging towards $h_i^k$ , the maximum optimization step of each feature dragging is set as 5. To enhance efficiency, we pause the optimization process if $L_{(i,k)}^{en}$ already falls below $0.5\cdot l$. The final termination signal is obtained by determining if the remaining distance $||h_i^{k}-t_i||_2 \le 2$. 

\subsection{Directional Editing}
If the optional binary mask is provided by users, the mask loss can be obtained as:
\vspace{-2mm}
\begin{equation}\label{eq:loss_mask}
 \mathcal{L}_{mask} = {\left\| ({F_0 - F})  \odot ( 1 - M)   \right\|_1},
\vspace{-2mm}
\end{equation}
where $F_0$ denotes the initial feature without any dragging, and $\odot$ is the element-wise multiplication. The total training loss can be expressed as:
\vspace{-2mm}
\begin{equation}\label{loss:total}
 \mathcal{L}_{total} = \mathcal{L}_{drag} + \gamma \cdot \mathcal{L}_{mask}.
\vspace{-2mm}
\end{equation}
\vspace{-2mm}
where $\gamma$ is the hyperparameter for loss balance.

\section{Experiments}
\label{sec:experiments}

Since the proposed feature dragging pipeline is constructed based on the feature map, thus it can be effortlessly implemented on StyleGAN2 models \cite{stylegan-2} and latent diffusion models\cite{stablediffusion} by extracting corresponding feature maps.

\subsection{Implementation Details}
Parameter $r $ in Eq. \ref{feature_aggregate} is set as 3, and parameter $\gamma $ in Eq. \ref{loss:total} is set as 10. 
For StyleGAN2 models, the feature map is extracted after the 6th block and the optimization for latent code is conducted in the extended $\mathcal{W}^+$ space\cite{image2stylegan}. We set $l=0.4$ and $d=4$ for elephant and lion models that are observed to likely perform larger movement in a single optimization step, and $l=0.3$ and $d=3$ for other StyleGAN2 models.  For diffusion models, following DragDiffusion\cite{dragdiffusion}, we fine-tune a LoRA \cite{lora} with rank of 16 on the UNet parameters for each image, which is used for both image inversion and dragging editing, and the feature map is extracted from the U-Net. We also replace the feature map with diffusion latent in Eq. \ref{eq:loss_mask} to keep consistent with DragDiffusion.
The parameters $l$ and $d$ are empirically set as 1 and 5 in diffusion models, respectively. To reflect the performance of different dragging pipelines themselves, FreeDrag and DragDiffusion utilize the same LoRA parameters for the same image.
To fully capture the potential of each method, the max step is set as 300 for all methods.

\subsection{ Dataset Construction }
\begin{figure}[]\centering
\includegraphics[width=0.85\linewidth]{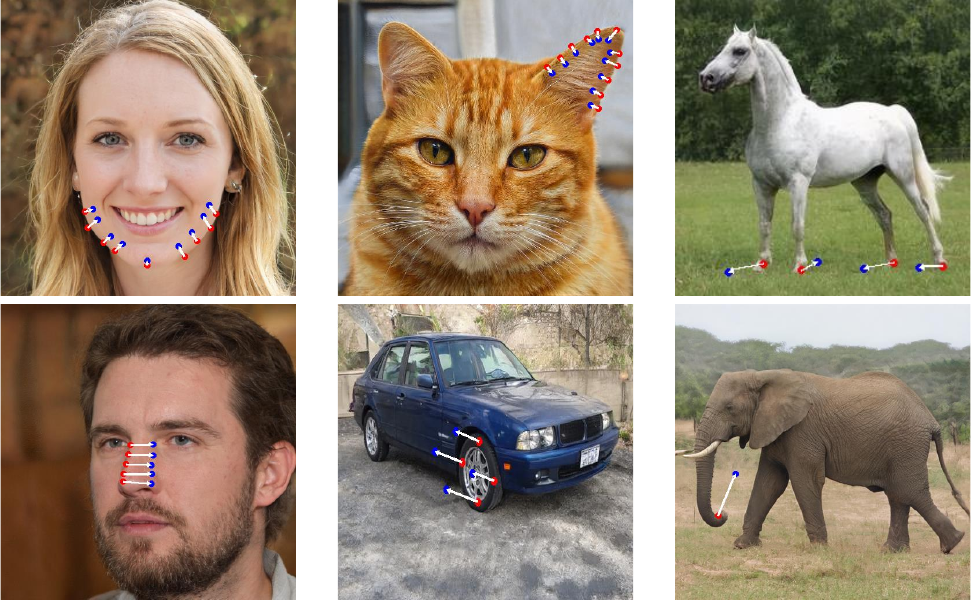}
\vspace{-3mm}
\caption{Several examples in the proposed FreeDragBench.}
\label{fig:dataset}
\vspace{-3mm}
\end{figure}

Since there is no public dataset to evaluate the drag-based editing in StyleGAN2, we propose FreeDragBench, which is the first dataset customized for GAN-based dragging editing. As presented in Table \ref{tab:dataset}, FreeDragBench consists of 600 images randomly generated by five different StyleGAN2 models, equipped with 2251 dragging instructions tailored for image content (including the editing in the pose, size, position, etc.), as shown in Fig. \ref{fig:dataset}. 
 
\begin{figure}[]\centering
\includegraphics[width=0.95\linewidth]{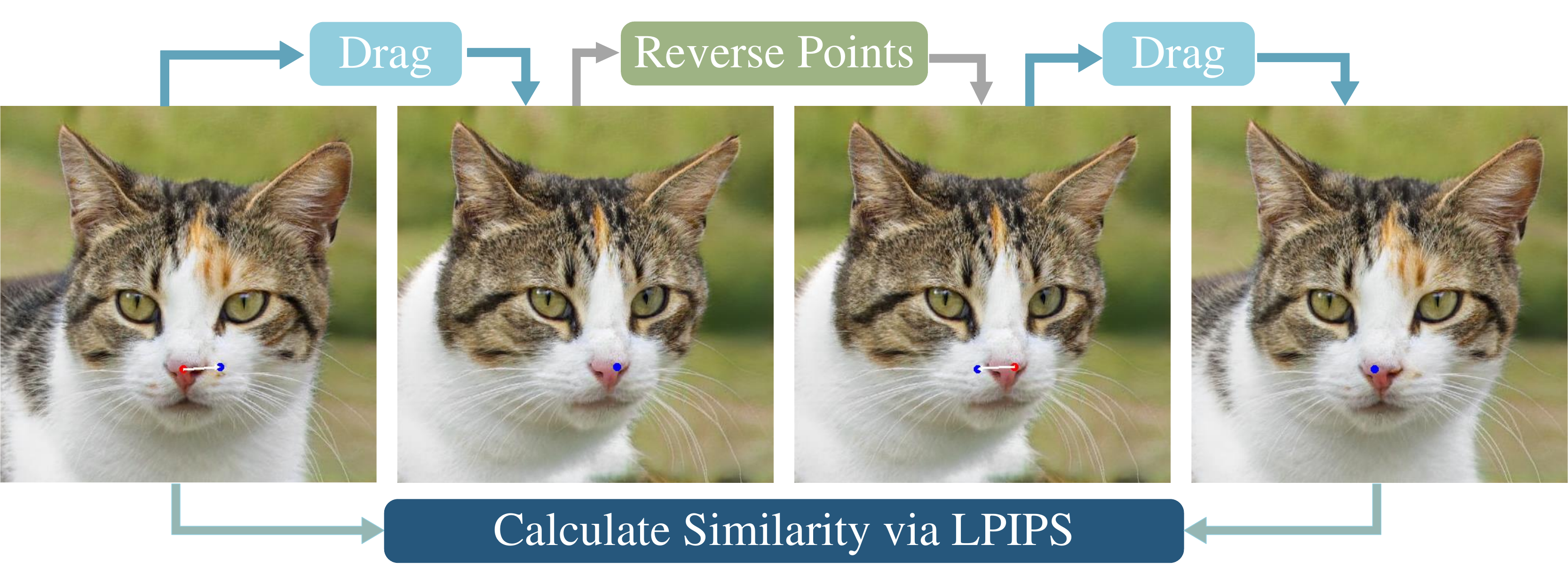}
\vspace{-4mm}
\caption{Visualization of the proposed CCSD metric. }
\label{fig:metric}
\vspace{-3mm}
\end{figure}

Furthermore, since the ground-truth corresponding to dragging instruction is not available, we propose a new metric to measure the accuracy of dragging editing, \emph{i.e.}, the Content Consistency under Symmetrical Dragging (CCSD). To be specific, as depicted Fig. \ref{fig:metric}, we reuse the reverse side of the original dragging instruction to construct a symmetrical dragging instruction pair and measure the content consistency under this symmetrical dragging instruction pair. To avoid penalizing stochastic elements with no effect on perception, LPIPS\cite{LPIPS} is used for similarity measurement. A low CCSD value requires accurate dragging in symmetrical editing, which could be used as an effective measurement metric in the absence of ground-truth. 

\subsection{Qualitative Evaluation}

\begin{figure*}[]\centering
\includegraphics[width=0.92\linewidth]{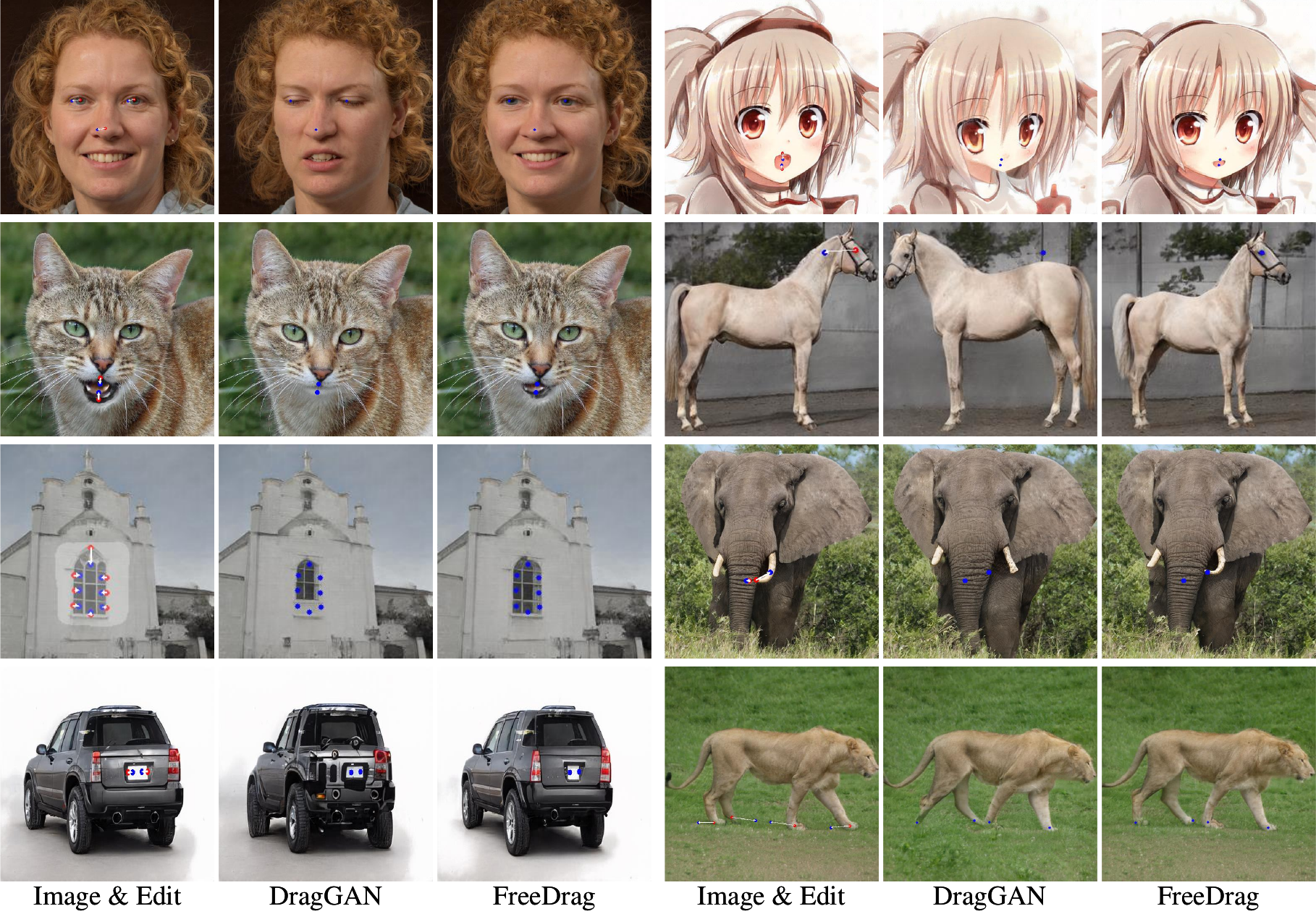}
\vspace{-2mm}
\caption{Demonstration of the edited results of FreeDrag and DragGAN\cite{dragondiffusion} in eight different StyleGAN2 models. 
}
\label{fig:comparison_gan}
\vspace{-3mm}
\end{figure*}

As depicted in Fig. \ref{fig:comparison_gan}, FreeDrag successfully avoids the abnormal disappearance of handle points (\emph{e.g.}, the vanished eyes in the human face, and the mouth of cartoon character and cat), showcasing its superiority in fine-detail editing. Meanwhile, FreeDrag achieves better stability against drastic content distortions (see the eye of the horse), steadily attaining the editing intent. Moreover, FreeDrag exhibits better robustness in handling similar points, resulting in reliable and precise dragging editing, as demonstrated in the examples of the third row. Additionally, FreeDrag effectively mitigates the potential misguidance during optimization steps, leading to more natural and coherent editing results, as observed in the last row in Fig. \ref{fig:comparison_gan}.

\begin{table}[]
\centering
\setlength{\tabcolsep}{1.5mm}
\begin{tabular}{c|ccccc}
\hline
Category & Face & Cat & Car & Horse & Elephant \\ \hline
Image number & 200 & 100 & 100 & 100 & 100 \\ \hline
Instruction number & 1068 & 406 & 337 & 227 & 213 \\ \hline
\end{tabular}
\vspace{-2mm}
\caption{Statistic of images and instructions of FreeDragBench.}
\label{tab:dataset}
\vspace{-3mm}
\end{table}

\begin{figure}[]\centering
\includegraphics[width=0.95\linewidth]{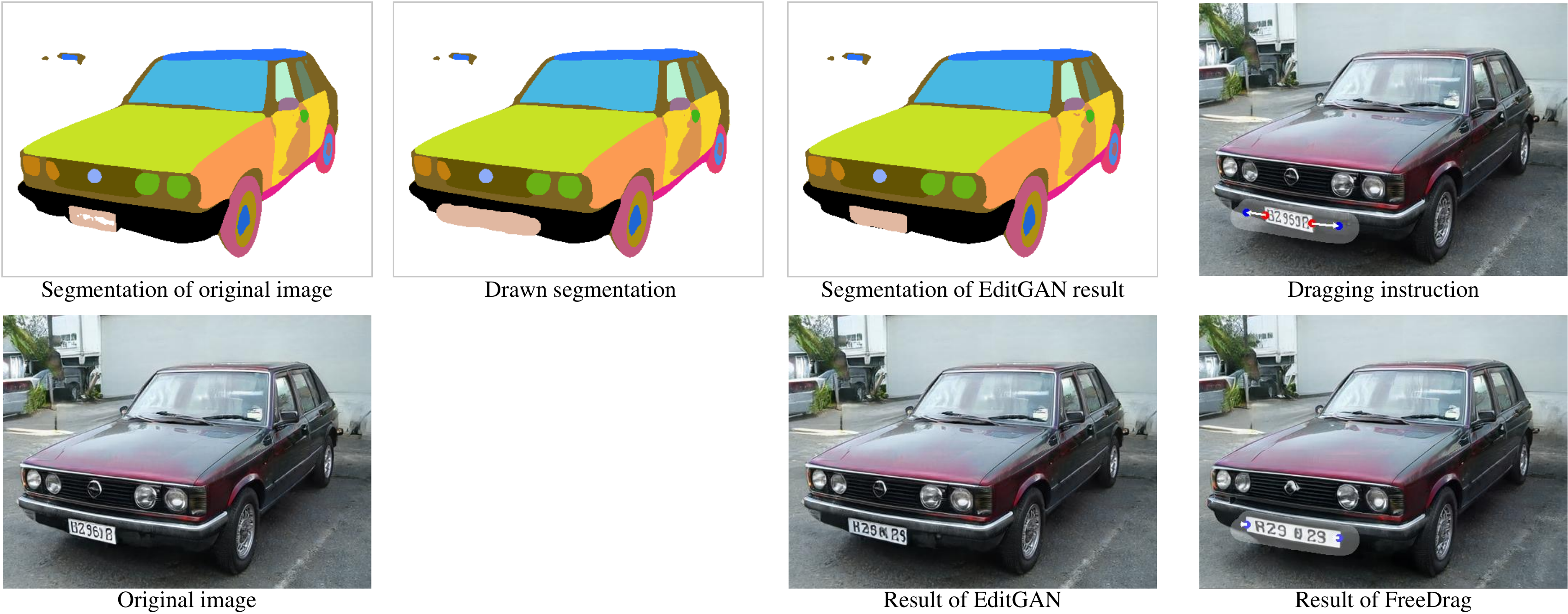}
\vspace{-4mm}
\caption{Comparison with EditGAN\cite{ling2021editgan} in editing accuracy. }
\label{fig:comparison_editgan}
\vspace{-6mm}
\end{figure}

\begin{table*}[]
\centering
\setlength{\tabcolsep}{2.3mm}
\begin{tabular}{cccccccccccc}
\hline
Category & \multicolumn{2}{c}{Face} & \multicolumn{2}{c}{Cat} & \multicolumn{2}{c}{Car} & \multicolumn{2}{c}{Horse} & \multicolumn{2}{c}{Elephant} & Time \\
Metric & FID & CCSD & FID & CCSD & FID & CCSD & FID & CCSD & FID & CCSD & Second \\ \hline
DragGAN\cite{pan2023drag} & 38.07 & 0.83 & 19.14 &0.56 & 36.36 & 0.73 & 21.90 & 1.20 & 11.17 &  1.19 &  8.26\\
FreeDrag & \textbf{29.50} & \textbf{0.35} & \textbf{15.67} & \textbf{0.23} & \textbf{33.50} & \textbf{0.37} & \textbf{21.18} & \textbf{0.68} & \textbf{10.86} & \textbf{0.82} & \textbf{2.74}\\ \hline
\end{tabular}
\vspace{-3mm}
\caption{Quantitative evaluation on FreeDragBench. A lower FID score indicates better fidelity in single dragging editing, while lower CCSD $(\times 10)$ scores imply higher accuracy in two symmetrical dragging editing. The time is calculated on Face category. }
\label{tab:comparsion_gan}
\vspace{-4mm}
\end{table*}

\begin{table}[h]
\centering
\setlength{\tabcolsep}{1mm}
\begin{tabular}{cccc}
\hline
DragBench     & MD $\downarrow$  & LPIPS $(\times 10)\downarrow$      & Time (Sec) $\downarrow$  \\ \hline
DragDiffusion\cite{dragdiffusion} & 38.76          & 1.38       & 71.77   \\ \hline
FreeDrag      & \textbf{33.49} & \textbf{1.02}  & \textbf{63.62} \\ \hline
\end{tabular}
\vspace{-3mm}
\caption{Quantitative evaluation on DragBench. The time consumption is computed on DragBench which only includes the dragging process because a fine-tuned LoRA can be used for multiple image editing with different dragging instructions.}
\label{tab:comparison_dif}
 \vspace{-3mm}
\end{table}

\begin{figure*}[]\centering
\includegraphics[width=0.92\linewidth]{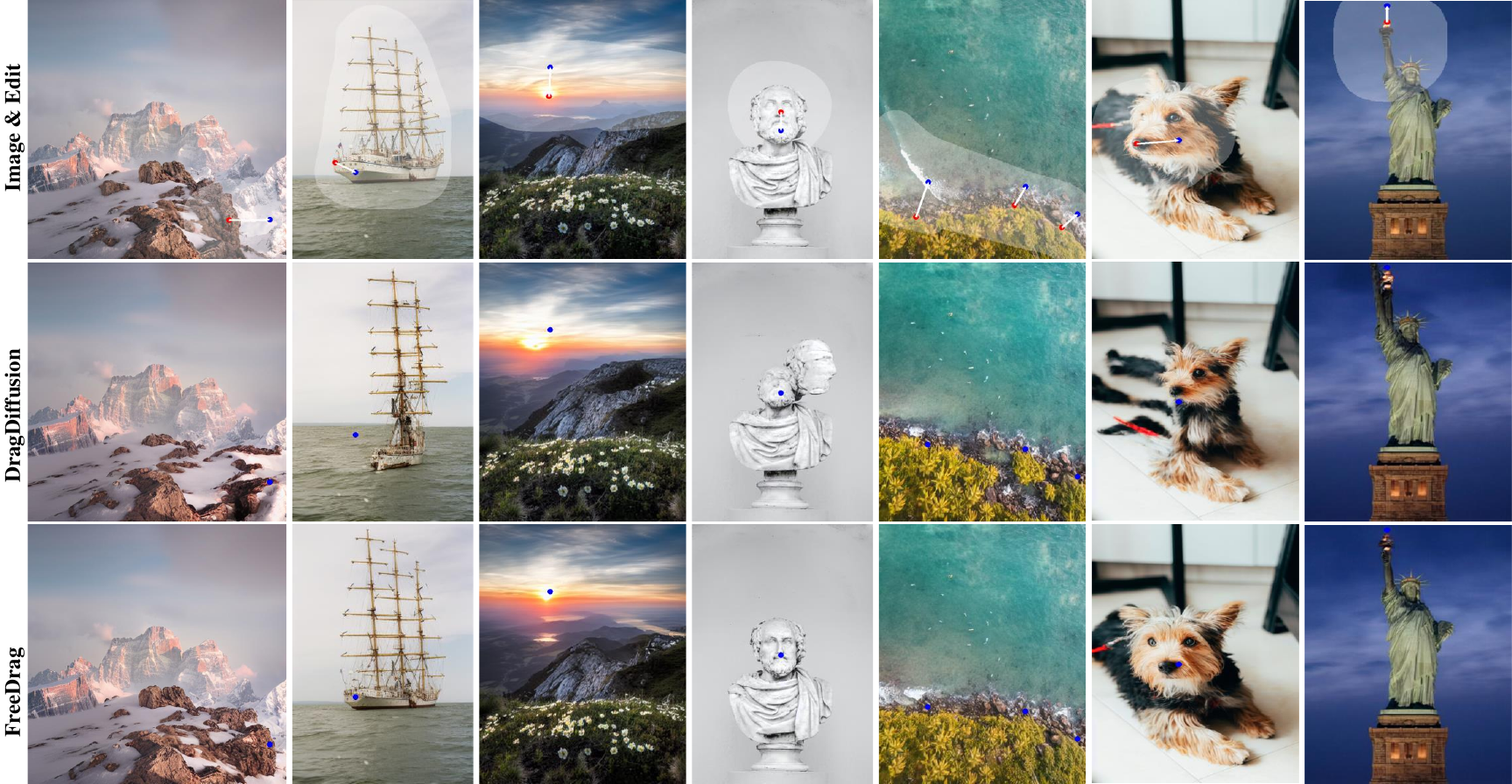}
\vspace{-3mm}
\caption{Demonstration of real image editing results of FreeDrag and DragDiffusion\cite{dragondiffusion}. }
\label{fig:comparison_diffusion}
\vspace{-2mm}
\end{figure*}

For image editing with the combination of diffusion models, FreeDrag also attains impressive performance. As shown in Fig. \ref{fig:comparison_diffusion}, FreeDrag outperforms DragDiffusion in both editing accuracy (see the examples from the first to third columns) and structure preservation (see the examples from the fourth to last columns), thus achieving superior quality of point-based dragging editing.

Additionally, we further conduct a comparison with EditGAN\cite{ling2021editgan}, which performs fine-grained editing by drawing object-level masks. As shown in Fig. \ref{fig:comparison_editgan}, FreeDrag better follows editing instructions.

\subsection{Quantitative Evaluation}
For quantitative evaluation, we implement comparison with DragGAN and DragDiffusion in FreeDragBench and DragBench\cite{dragdiffusion}, respectively. Specifically, for the comparison in FreeDragBench, we use FID and the proposed CCSD to evaluate the image quality and editing accuracy, respectively. For DragBench that owns images with varying resolution, we follow the setting in DragDiffusion\cite{dragdiffusion}, \textit{i.e.}, Mean Distance (MD) for dragging accuracy measurement and LPIPS \cite{LPIPS} for image fidelity evaluation. The mean distance is obtained by calculating the corresponding relationship of points between the original image and the edited image based on DIFT\cite{DIFT}. 

As presented in Table \ref{tab:comparsion_gan}, FreeDrag consistently attains higher scores in all categories, which further validates its superiority in achieving precise dragging editing and better image fidelity preservation. Moreover, it can be observed that FreeDrag gains significant improvement in time consumption, which can be attributed to that the proposed line search effectively alleviates the interference of similar points and thus successfully avoids unrewarding dragging steps, allowing for higher efficiency.

\begin{figure}[h]\centering
\includegraphics[width=0.8\linewidth]{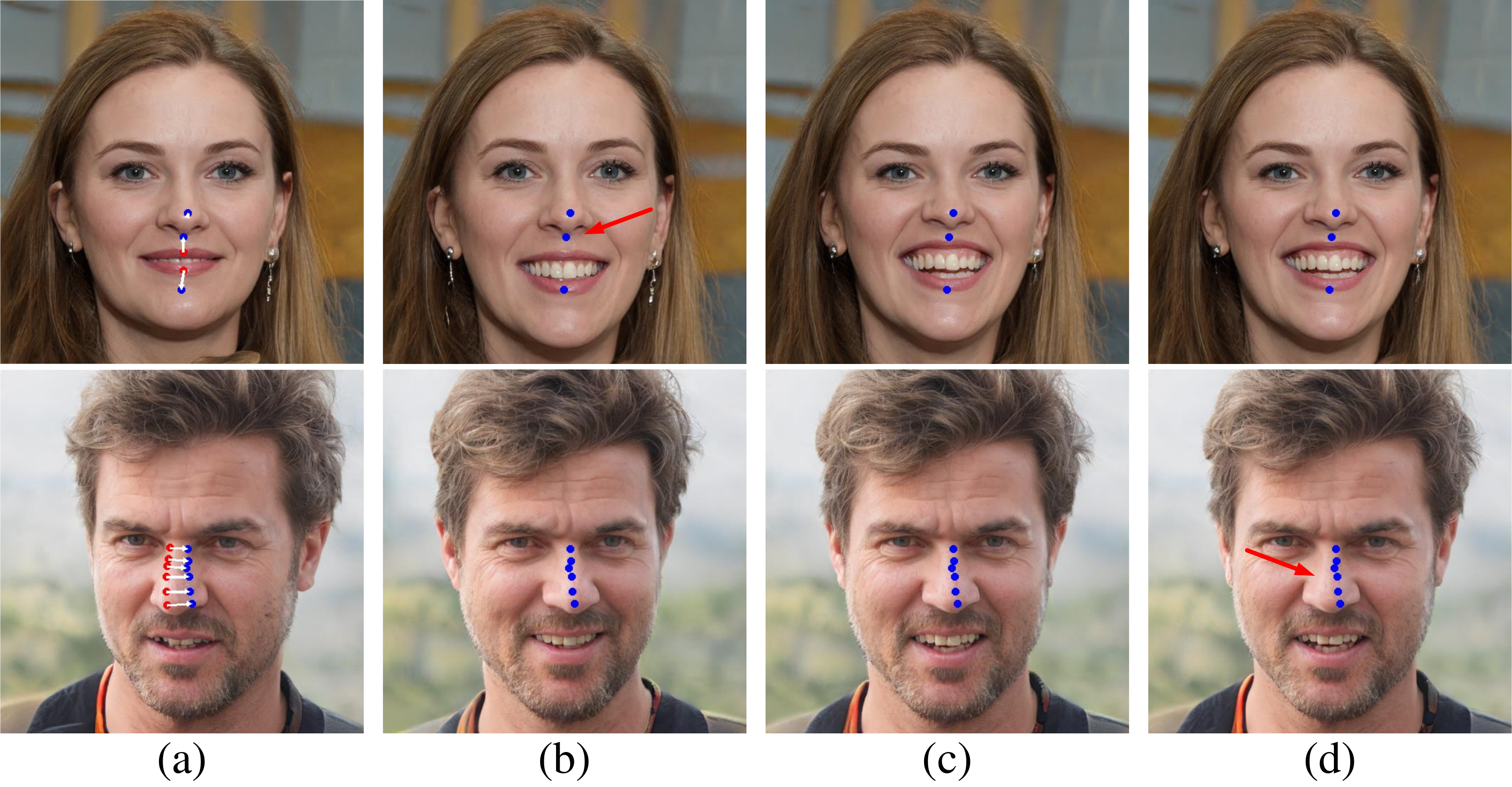}
\vspace{-2mm}
\caption{The edited results by using different parameters. (a) Original images with dragging instructions. (b) Edited results with $\left \{  l=0.15, d=1.5\right \} $. (c) Edited results with $\left \{  l=0.3, d=3\right \} $. (d) Edited results with $\left \{  l=0.45, d=4.5\right \} $.}
\label{fig:ablation_1}
\vspace{-2mm}
\end{figure}

\begin{figure}[h]\centering
\includegraphics[width=0.8\linewidth]{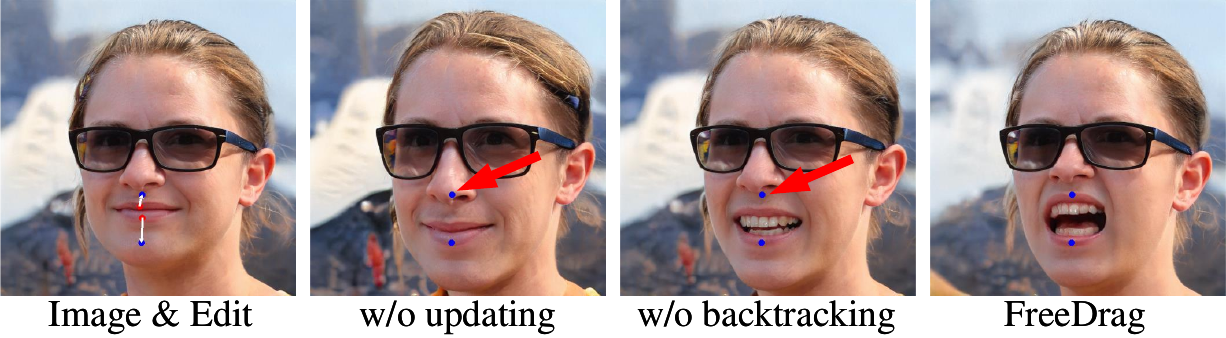}
\vspace{-2mm}
\caption{Illustration of the effect of adaptive updating strategy in template feature and backtracking mechanism in line search.}
\label{fig:ablation_2}
\vspace{-4mm}
\end{figure}

\begin{table}[]
\centering
\setlength{\tabcolsep}{1mm}
\begin{tabular}{cccc}
\hline
 Metric    & w/o updating  & w/o backtracking     & Ours  \\ \hline
 CCSD $(\times 10)$ &0.82          & 0.52      & \textbf{0.35}   \\ \hline
\end{tabular}
\vspace{-2mm}
\caption{Quantitative ablation on human face model.}
\label{tab:ablation}
 \vspace{-6mm}
\end{table}

For the quantitative evaluation in diffusion models, we utilize the public DragBench dataset \cite{dragdiffusion} that is customized for diffusion-based dragging evaluation. The results of DragDiffusion and FreeDrag are presented in Table. \ref{tab:comparison_dif}. It is observed that FreeDrag outperforms DragDiffusion with higher dragging accuracy and lower time-consumption, implying a promising potential for versatile applications.

\subsection{Ablation Study}
The parameters $l$ and $d$ determine the initial feature discrepancy and maximum single movement distance, thus controlling the style of total dragging editing.  Specifically, a too small $l$ or $d$ implies a more conservative editing strategy, which prefers small motion and refuses large updating scale, thus failing to reach the target point in limited optimization steps, as shown in Fig. \ref{fig:ablation_1}(b). In contrast, a too large $l$ or $d$ means a more impulsive editing strategy, which appears to accept large updating scale and larger movement distance and thus increases the risk of coarse feature updating, resulting in damage to editing accuracy, as can be observed in Fig. \ref{fig:ablation_1}(d).

Furthermore, we assign $\lambda=0$ in Eq. \ref{eq:adaptive_updating} to obtain a stationary template feature to evaluate the effect of adaptive updating strategy and adopt Eq. \ref{eq.localization} rather than Eq. \ref{eq.whole_localization} to evaluate the effect of backtracking mechanism. As can be observed in Fig. \ref{fig:ablation_2}, both of them play necessary roles for better editing quality. The quantitative ablation in Table \ref{tab:ablation} also validates their necessity.

\subsection{Limitation}

\begin{figure}[h]\centering
\includegraphics[width=0.95\linewidth]{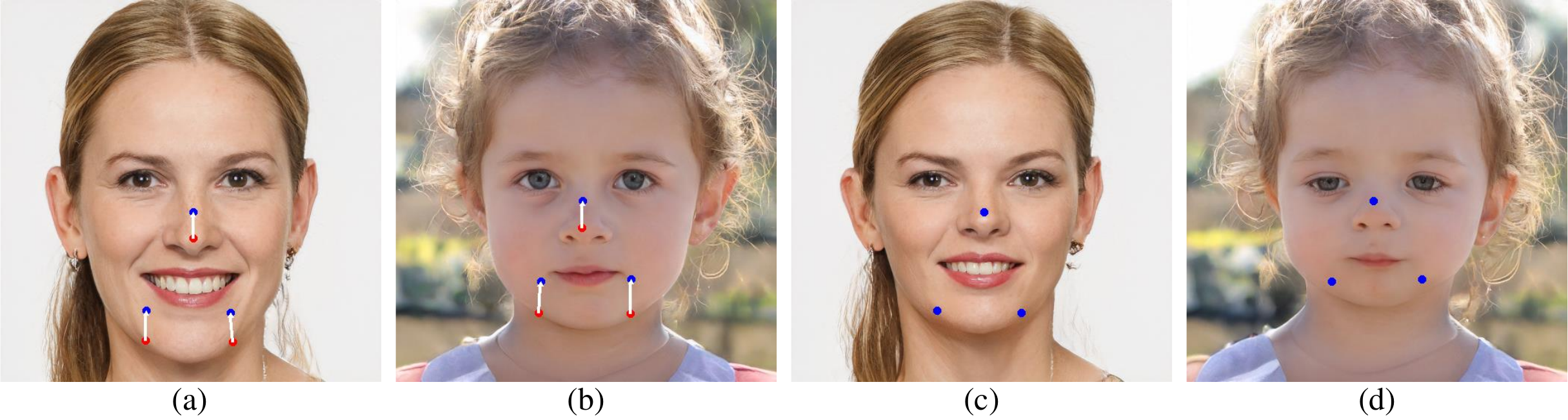}
\vspace{-2mm}
\caption{Limitations of 2D dragging instructions. (a)-(b): Dragging instructions. (c)-(d): Edited results.}
\label{fig:limitation}
\vspace{-4mm}
\end{figure}

Freedrag achieves image manipulation with a reliance on the 2D dragging instructions, which may be limited in certain 3D editing scenarios due to potential ambiguity. As can be observed in Fig. \ref{fig:limitation}, an intention of ``head up” might result in a ``resized shape” instead.

\section{Conclusion}
\label{sec:conclusion}

In this work, we propose FreeDrag, a novel feature dragging framework for reliable point-based image editing. By incorporating an adaptive template feature, FreeDrag allows for flexible control in the scale of each feature updating, which contributes to stronger stability under drastic content change, resulting in a better immunity against point missing.
Meanwhile, the established line search with backtracking effectively mitigates the misguidance caused by similar points and allows timely adjustment for motion plan by effectively discriminating abnormal motion, leading to reliable and continuous movements towards the final target point.
Extensive experiments demonstrate the reliability of FreeDrag in precise semantic dragging and stable structure preservation, indicating superior editing quality.

\noindent
\textbf{Acknowledgement.} This work is supported in part by the Postdoctoral Fellowship Program of CPSF GZB20230713, in part by the Anhui Provincial Key Research and Development Plan  202304a05020072, in part by the Fundamental Research Funds for the Central Universities  WK2090000065, and in part by the Anhui Provincial Natural Science Foundation 2308085QF226. This work is partially supported by the National Key R\&D Program of China (2022ZD0160201), and Shanghai Artificial Intelligence Laboratory.

{
    \small
    \bibliographystyle{ieeenat_fullname}
    \bibliography{main}
}

\newpage
\begin{figure*}[t]
  \centering
  \includegraphics[width=1\linewidth]{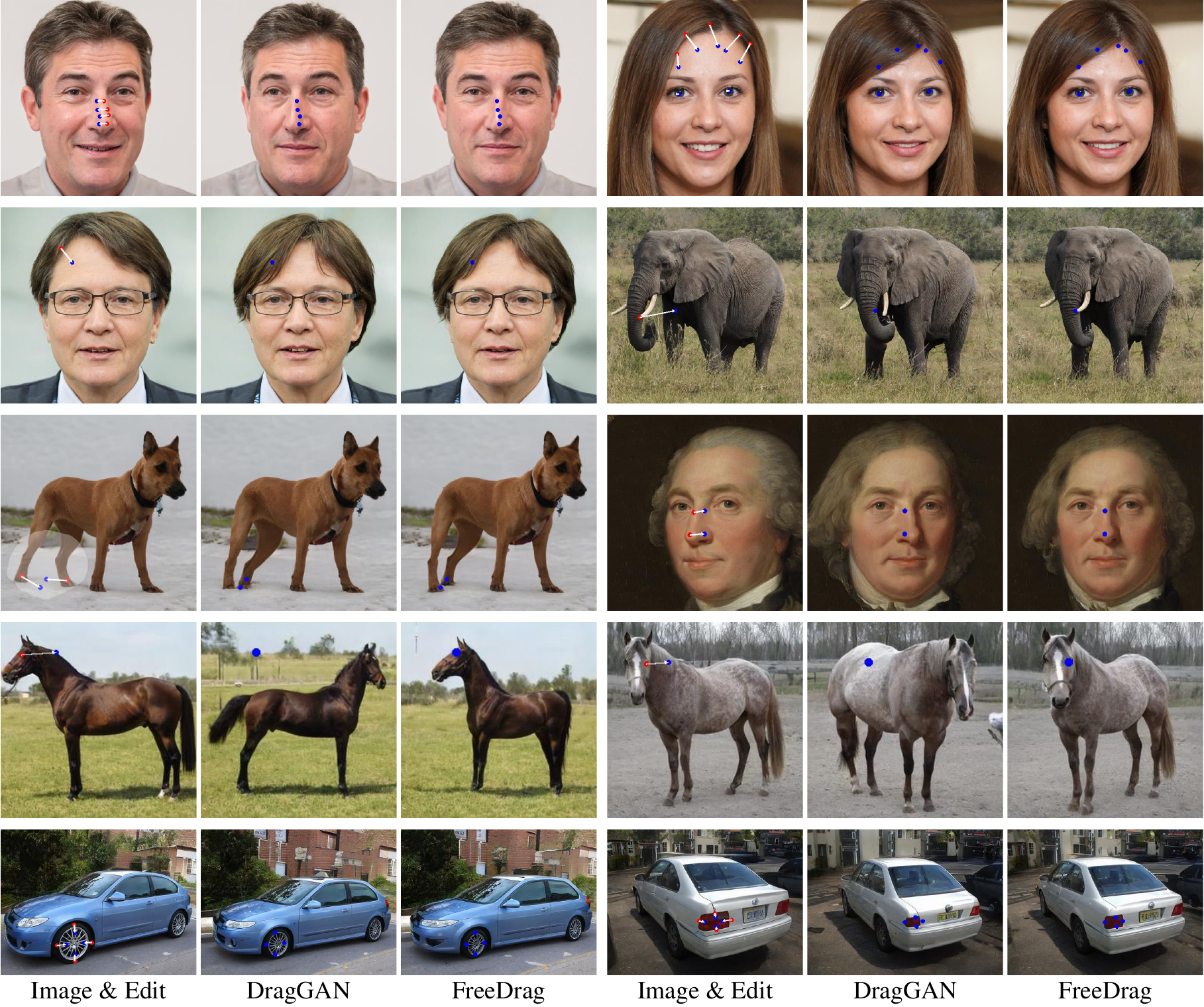}
   \caption{Visual comparison in GAN-based dragging results. FreeDrag achieves more precise point-based dragging editing.}
   \label{fig:sup_comparison_gan-1}
\end{figure*}

\begin{figure*}[t]
  \centering
  \includegraphics[width=0.85\linewidth]{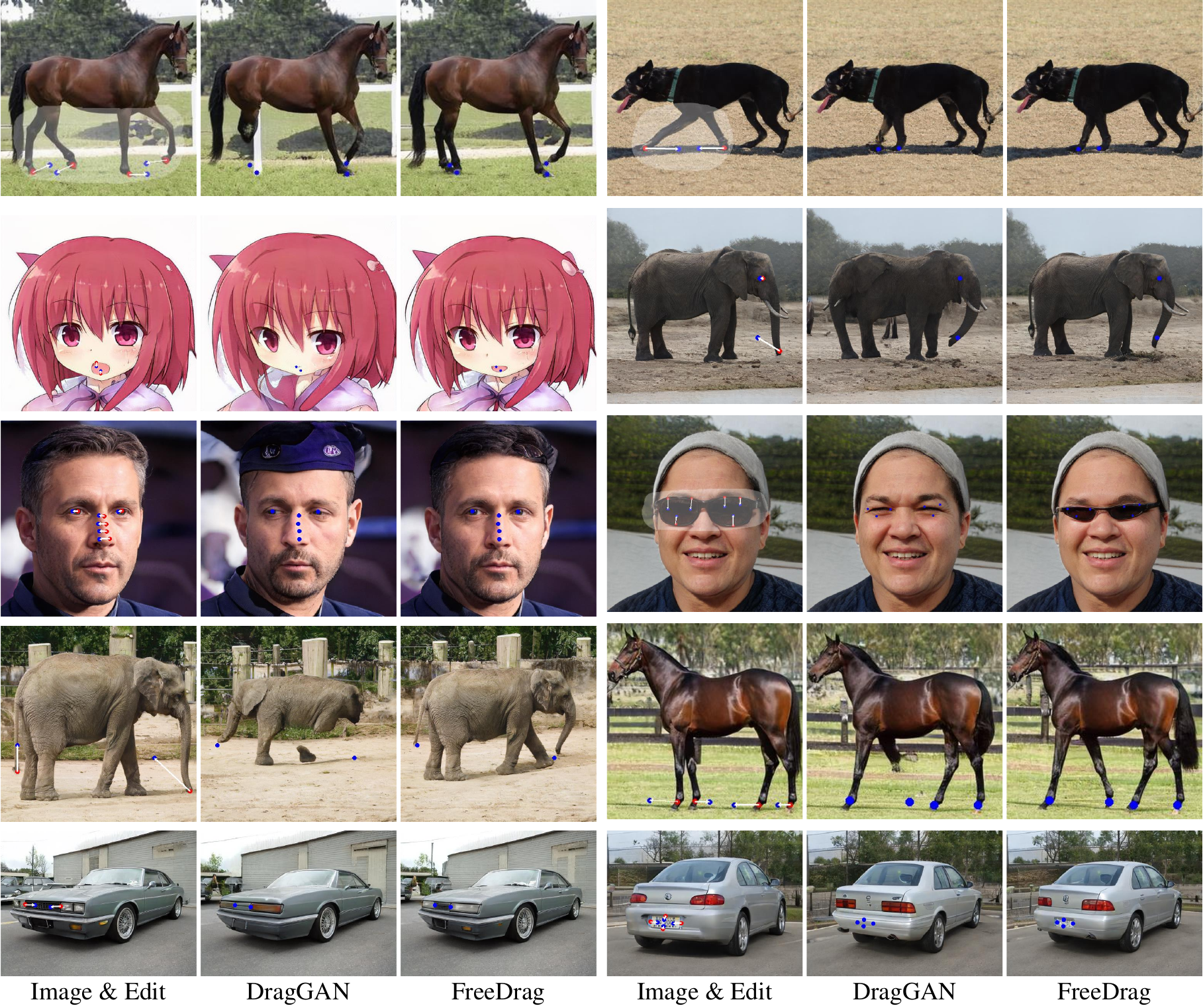}
   \caption{Visual comparison in GAN-based dragging results. FreeDrag achieves superior preservation of original structure.}
   \vspace{-10mm}
   \label{fig:sup_comparison_gan-2}
   \vspace{-2mm}
\end{figure*}

\begin{figure*}[t]
  \centering
  \includegraphics[width=0.9\linewidth]{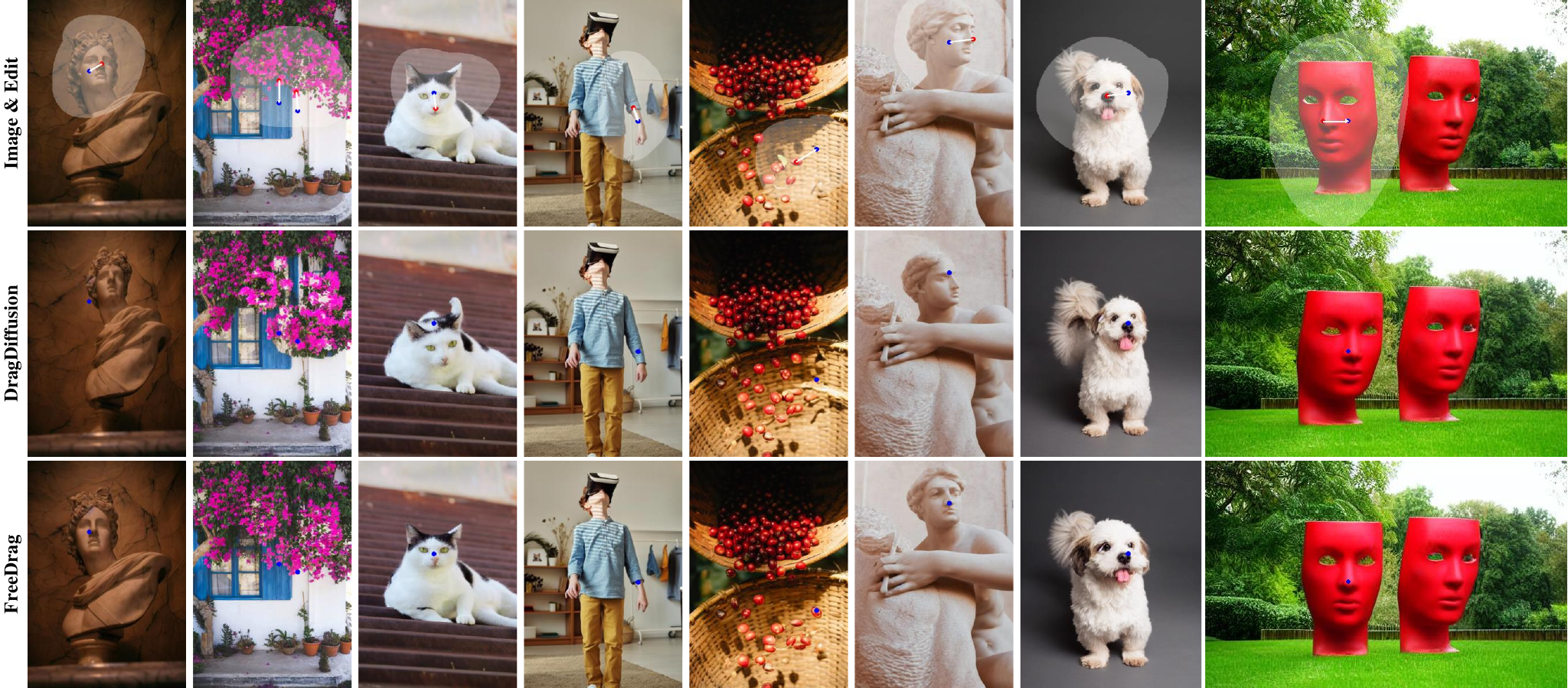}
   \caption{Visual comparison in real image editing. FreeDrag outperforms DragDiffusion in precise dragging and structure preservation.}
   \vspace{-3mm}
   \label{fig:sup_comparison_diffusion}
\end{figure*}

\end{document}